\documentclass[final]{cvpr}

\usepackage{cuted}
\usepackage{capt-of}
\usepackage{soul,color}
\usepackage{times}
\usepackage{epsfig}
\usepackage{graphicx}
\usepackage{amsmath}
\usepackage{amssymb}
\usepackage{subcaption}
\usepackage{multirow}
\usepackage[ruled,vlined]{algorithm2e}
\usepackage{enumitem}
\usepackage{floatrow}

\SetCommentSty{mycommfont}

\usepackage[pagebackref=true,breaklinks=true,colorlinks,bookmarks=false]{hyperref}

\def\cvprPaperID{29} 
\def\confYear{CVPR 2021}

\begin{document}
\setlength{\abovedisplayskip}{2pt}
\setlength{\belowdisplayskip}{2pt}
\title{Tracking Grow-Finish Pigs Across Large Pens Using Multiple Cameras}

\author{Aniket Shirke \hspace{1mm} Aziz Saifuddin \hspace{1mm} Achleshwar Luthra \hspace{1mm} Jiangong Li \hspace{1mm} Tawni Williams \hspace{1mm} Xiaodan Hu \\ \hspace{1mm}  Aneesh Kotnana \hspace{1mm} Okan Kocabalkanli \hspace{1mm} Narendra Ahuja \hspace{1mm} Angela Green-Miller \hspace{1mm} Isabella Condotta \\ \hspace{1mm} Ryan N. Dilger \hspace{1mm} Matthew Caesar \\
University of Illinois, Urbana-Champaign\\
{\tt\small anikets@illinois.edu}
}


\maketitle

\begin{abstract}
Increasing demand for meat products combined with farm labor shortages has resulted in a need to develop new real-time solutions to monitor animals effectively. Significant progress has been made in continuously locating individual pigs using tracking-by-detection methods.  However, these methods fail for oblong pens because a single fixed camera does not cover the entire floor at adequate resolution. We address this problem by using multiple cameras, placed such that the visual fields of adjacent cameras overlap, and together they span the entire floor. Avoiding breaks in tracking requires inter-camera handover when a pig crosses from one camera’s view into that of an adjacent camera. We identify the adjacent camera and the shared pig location on the floor at the handover time using inter-view homography. Our experiments involve two grow-finish pens, housing 16-17 pigs each, and three RGB cameras. Our algorithm first detects pigs using a deep learning-based object detection model (YOLO) and creates their local tracking IDs using a multi-object tracking algorithm (DeepSORT). We then use inter-camera shared locations to match multiple views and generate a global ID for each pig that holds throughout tracking. To evaluate our approach, we provide five 2-minutes long video sequences with fully annotated global identities. We track pigs in a single camera view with a Multi-Object Tracking Accuracy and Precision of 65.0\% and 54.3\% respectively and achieve a Camera Handover Accuracy of 74.0\%. We open-source our code and annotated dataset at \url{https://github.com/AIFARMS/multi-camera-pig-tracking}
\end{abstract}

\begin{figure*}[h]
    \begin{subfigure}{.33\textwidth}
      \centering
      \includegraphics[width=0.99\linewidth,height=2.3cm]{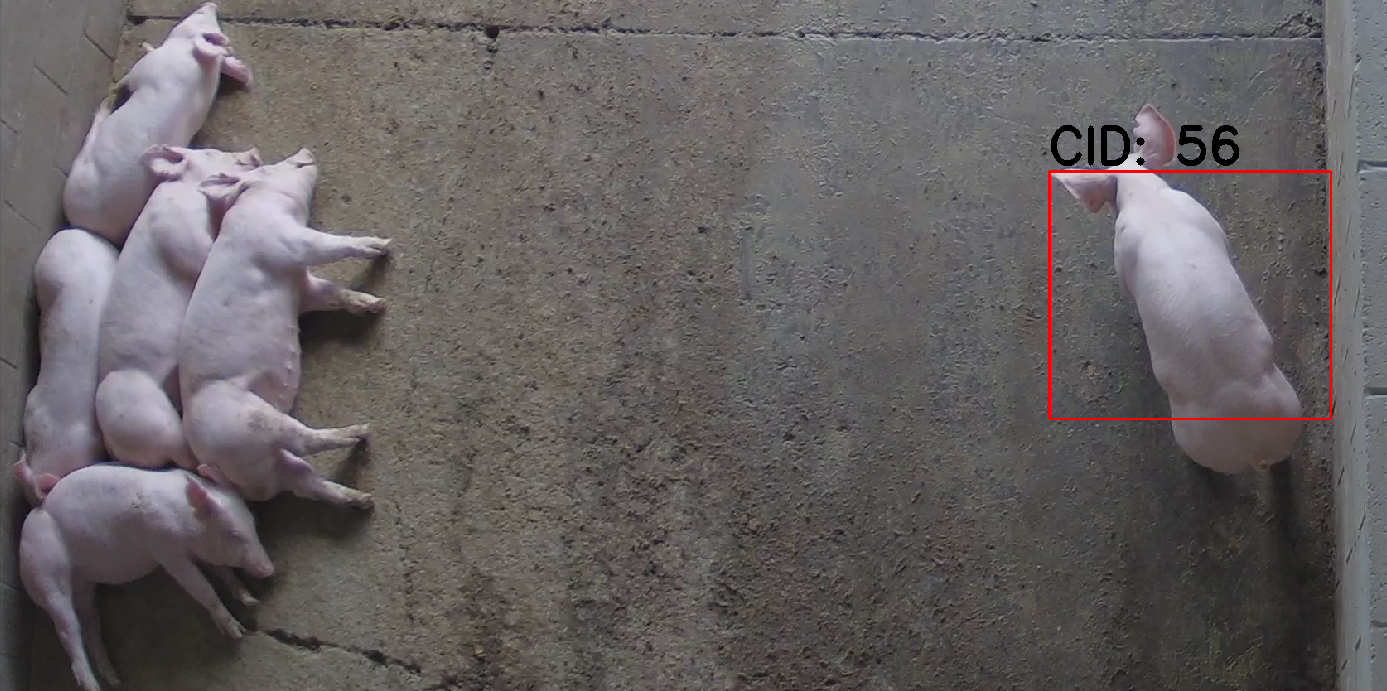}
      \caption{Track detected in \textit{Ceiling} view}
      \label{fig:ceiling}
    \end{subfigure}%
    \begin{subfigure}{.33\textwidth}
      \centering
      \includegraphics[width=0.99\linewidth,height=2.3cm]{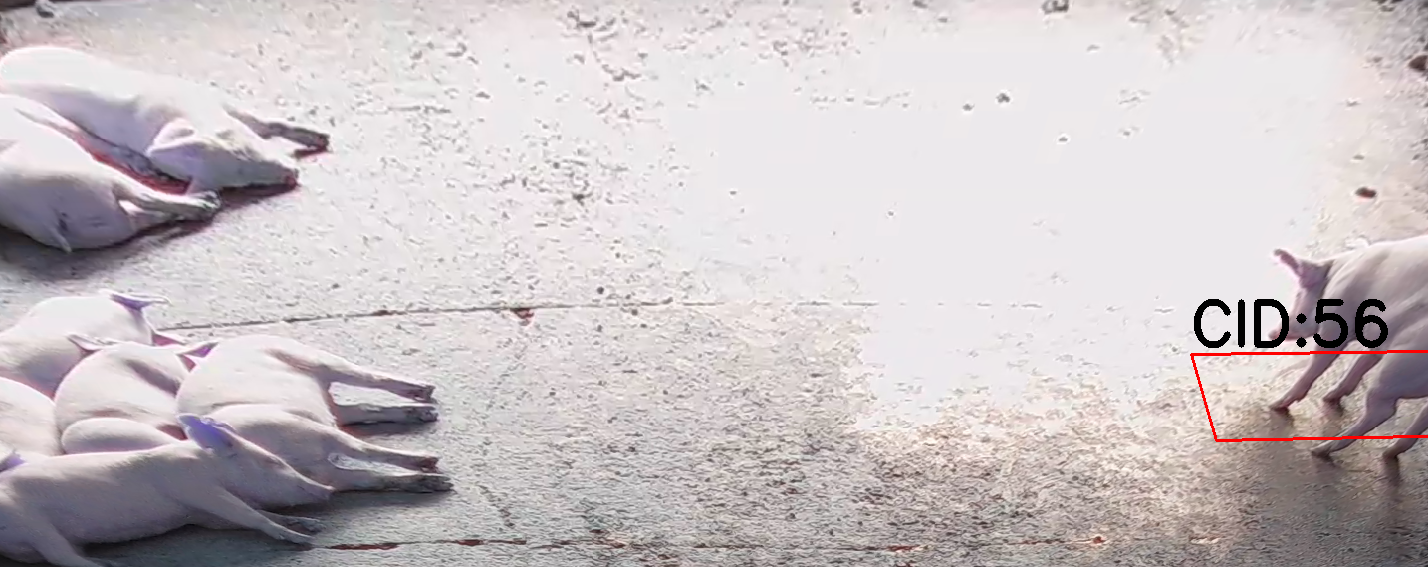}
      \caption{Perspective warp using homography}
      \label{fig:homography}
    \end{subfigure}%
    \begin{subfigure}{.33\textwidth}
      \centering
      \includegraphics[width=0.99\linewidth,height=2.3cm]{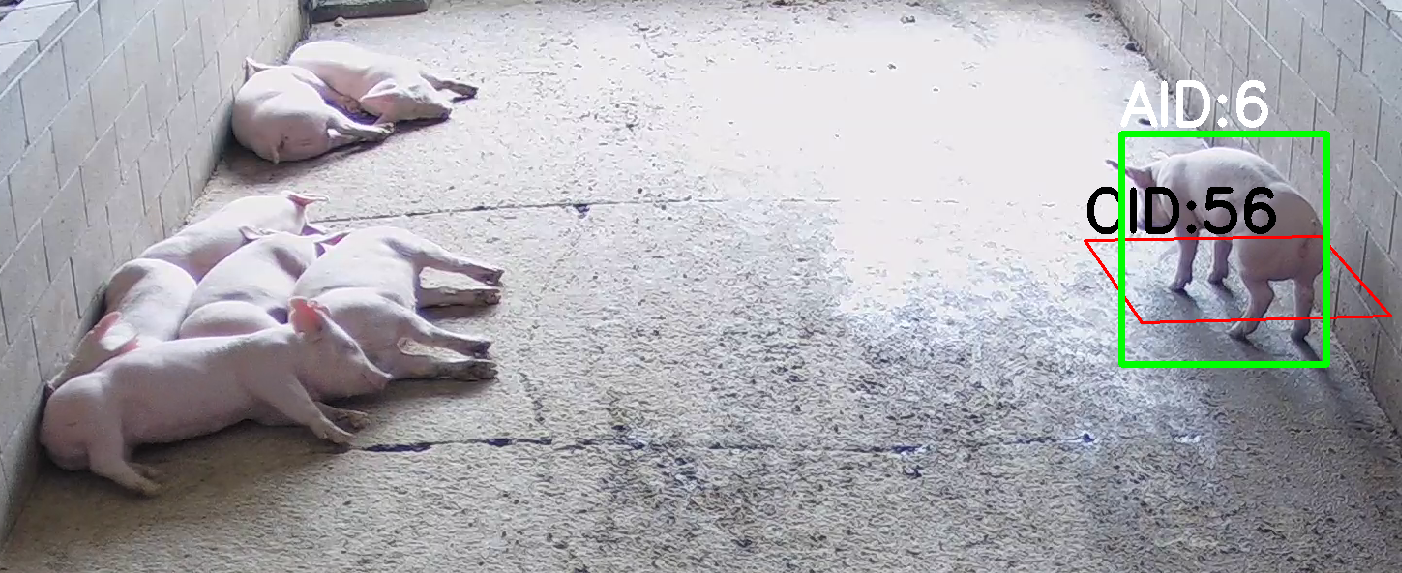}
      \caption{Matching track in \textit{Angled} view}
      \label{fig:angled}
    \end{subfigure}%
\caption{Simplified depiction of the Multi-Camera Tracking algorithm}
\vspace{-0.2cm}
\end{figure*}

\section{Introduction}

Pigs are one of the most commonly raised livestock animals in the world, forming a primary protein source for millions of people across numerous cultures and geographical regions \cite{shahbandeh}. The increase in demand for high-quality protein has resulted from growing populations combined with increasing incomes across the globe.

Health is a contributor to efficient production, and ideally could be best done by monitoring them individually. Behavior has been well-established as a strong indicator of pig health\cite{pig-health}. Manual monitoring of individual behavior in a commercial livestock farm is not practical or sustainable with existing farm staffing and workflow \cite{swan}. Each animal typically receives no more than a few seconds of observation time on an average day \cite{wean-to-finish-manual}. 

Monitoring pigs automatically requires tracking each animal and identifying and interpreting its behavior. This presents several key challenges. Pens often contain a large number of pigs, that are similar in appearance and thus difficult to visually distinguish from each other. Moreover, pens are often in buildings with low ceilings or otherwise congested areas, making it difficult to see the entire pen from a single vantage point, requiring monitoring across different views, from different vantage points.

To address this challenge, we develop a system to monitor pigs using multiple cameras with adjacent cameras having overlapping fields of view. We detect and track pigs in a given view using state-of-the-art object detection and tracking models. We then use homography between adjacent cameras to identify a pig in overlapping views, transitively match different camera views of the same pig, and assign a global identity to each pig, thereby achieving global tracking. In order to validate our methods, we have monitored two grow-finish pens over a period of eight weeks using a set of appropriately chosen and placed cameras.

Our primary contributions in this paper can be summarized as follows:
\begin{itemize}[noitemsep,topsep=0pt]
    \item We present a multi-camera pig tracking system for identifying each pig across all camera views. We are not aware of any other such system.
    \item We present a multi-camera, multi-pen dataset containing videos of pigs in multiple pens captured from multiple viewpoints and make it publicly available. To the best of our knowledge, our dataset is the first of its kind.
\end{itemize}

\section{Related Work}
Individual tracking of animals in group housing is a demanding task. Uninterrupted tracking poses significant challenges when there is a lack of discernible differences in the physical characteristics of the animals. To address these challenges, tracking-by-detection methods have been proposed. \cite{t2020long} casts detection as a segmentation task. The four semantic parts of pig (ears, shoulder, and tail) are detected and tracked using a Fully Convolutional Network and the Hungarian algorithm. 
In \cite{seo2020embeddedpigdet}, a TinyYOLO \cite{yolo} architecture is employed to detect pigs from infrared videos, with emphasis on execution speed as the target platform is an embedded device. \cite{sa2019fast} proposes a method to detect pigs under various illumination conditions, by combining information from depth and infrared images using spatio-temporal interpolation. Similarly, in \cite{brunger2020panoptic} the bounding boxes are replaced with ellipses, which are detected through a segmentation network. \cite{zhang2019automatic} uses an SSD \cite{ssd} architecture coupled with the MOSSE \cite{mosse} algorithm to perform animal tracking. 

Prior work in multi-camera, multi-target tracking typically assumes that the target has distinguishing features, which can be used for re-identification across multiple views \cite{haar} \cite{Ristani_2018_CVPR}. But this has not led to a similar ability to track pigs because of the lack of such visual features on pigs. To the best of our knowledge, our work is the first application of multi-camera tracking to pigs. Moreover, the only few publicly available datasets for pig monitoring consist of a single-camera view (\cite{t2020long}, \cite{bergamini2021extracting}, \cite{riekert2020automatically}), with no open-source implementations. To address this need, we collected and open-source our multi-camera view dataset, along with our implementation.
\section{Methods}

\begin{table*}
\begin{center}
 \begin{tabular}{|*{9}{c|}} 
    \hline
    \multirow{2}{*}{Type} & \multicolumn{7}{c|}{Local Tracking} & \multicolumn{1}{c|}{Global Tracking} \\\cline{2-9}
                        & \multicolumn{1}{c|}{IDF1} & \multicolumn{1}{c|}{IDP} & \multicolumn{1}{c|}{IDR} & \multicolumn{1}{c|}{Recall} & \multicolumn{1}{c|}{Precision} & \multicolumn{1}{c|}{MOTA} & \multicolumn{1}{c|}{MOTP} & \multicolumn{1}{c|}{CHA} \\
    \hline
    Day & 66.1 & 61.8 & 71.0  & 98.0  & 85.4  & 80.6\%  & 61.3\%  & 92.4\% \\
    Night & 53.2 & 49.9 & 56.9 & 85.6 & 75.1 & 55.2\% & 44.5\% & 46.3\% \\
    \hline
    Overall & 58.2 & 54.5 & 62.3 & 90.4 & 79.0 & 65.0\% & 54.3\% & 74.0\% \\
    \hline
\end{tabular}
\end{center}
\caption{Local and Global Tracking Metrics}\label{tab:metrics}
\end{table*}

\subsection{Detection and Local Tracking}
For each camera view, pig detection is achieved by using the state-of-the-art YOLOv4 \cite{bochkovskiy2020yolov4} model. In order to generate tracking IDs in a single camera view, the detections provided by YOLOv4 are tracked using DeepSORT \cite{deepsort}. The following subsections describe how local tracking IDs are processed to obtain global tracking IDs for pigs.

\subsection{Homography Estimation}
Any two images of the same planar surface are related by a homography. In our case, the planar surface is the pen floor, observed from a \textit{Ceiling} view and an \textit{Angled} view, as depicted in Figures \ref{fig:ceiling} and \ref{fig:angled}, respectively. 
Assuming that the \textit{Angled} view is aligned with the world coordinate axes and has no translation, a 3-D point $P$ is mapped to the 2-D pixel point $p_{angled}$ in the \textit{Angled} view using the following relation:
\begin{align}
\lambda_1 p_{angled} = K [I | 0] P
\end{align}
where $K$ is the intrinsic parameter matrix, $I$ is a 3x3 identity matrix, and $p_{angled}$ and $P$ are homogeneous coordinates defined as follows:
\begin{align}
p_{angled} = 
\begin{bmatrix}
x_{image} \\
y_{image} \\
1  
\end{bmatrix} 
P = 
\begin{bmatrix}
X_{world} \\
Y_{world} \\
Z_{world} \\
1  
\end{bmatrix} 
\end{align}
Similarly, the same 3-D point $P$ is mapped to the 2-D pixel point $p_{ceiling}$ in the \textit{Ceiling} view using the relation:
\begin{align}
\lambda_2 p_{ceiling} = K [R | t] P
\end{align}
where $R$ and $t$ is the relative rotation and translation of the \textit{Ceiling} view with respect to the \textit{Angled} view. Note that $\lambda_1$ and $\lambda_2$ are free parameters as multiple 3-D points can get mapped to the same 2-D point in the image due to single-view ambiguity. 
Our task is to estimate a homography $H(.)$ such that
\begin{align}x_{angled} = H(x_{ceiling})\end{align}

In an ideal scenario, having many 3D-2D correspondences between the floor pen and the corresponding image views can help in obtaining the homography accurately. Obtaining those correspondences is challenging at a farm due to the large numbers of pigs occluding the floor. In the absence of such correspondences, there are multiple methods that can be potentially used to estimate homography between two camera views. We adopt the following method to estimate the homography $
$:
\begin{enumerate}[noitemsep,topsep=-1pt]
    \item Using a perspective transformation, the \textit{Ceiling} view is first transformed into a top-down view parallel to the pen floor. The homography between these two views is denoted as $H_\text{ceiling$\rightarrow$top-ceiling}(.)$
    \item The \textit{Angled} view is then transformed into a top-down view using a similar perspective transformation. The homography between these two views is denoted as $H_\text{angled$\rightarrow$top-angled}(.)$.
    \item Since both the top-down views are parallel to the pen floor, key points from the overlapping regions of both these views are matched. These keys points can be detected as well as matched automatically. But in this paper, we perform these steps manually, by inspection, and estimate a homography using RANSAC. The homography is denoted as $H_\text{top-ceiling$\rightarrow$top-angled}(.)$.
\end{enumerate}
Thus, a pixel point $p_{ceiling}$ in the \textit{Ceiling} view can be mapped to a point $p_{estimated}$ in the \textit{Angled} view as follows:
\begin{align}
\begin{split}
p_{estimated} = {}& H_\text{angled$\rightarrow$top-angled}^{-1}(H_\text{top-ceiling$\rightarrow$top-angled}\\
&  (H_\text{ceiling$\rightarrow$top-ceiling}(p_{ceiling}))) \\
 = & H_\text{ceiling$\rightarrow$angled}(p_{ceiling})
\end{split}
\end{align}

\subsection{Track Aligning Algorithm}
\label{sec:global-tracking}

\begin{algorithm}[h]
\SetAlgoLined
\KwData{Set of $c$ and $a$ local pig IDs and bounding boxes in \textit{Ceiling} view and \textit{Angled} view respectively: \textit{ceiling\textunderscore tracks, angled\textunderscore tracks}}
\KwResult{Dictionary of matches between ceiling\textunderscore ids and angled\textunderscore ids: \textit{matches}}
matrix = [] \tcp*[l]{size of the matrix will be $c$x$a$}

\For{(ceiling\textunderscore id, bbox) \textbf{in} transformed\textunderscore ceiling\textunderscore tracks}{
matrix.append(intersection(bbox, angled\textunderscore tracks))\;
}
\tcc{Greedily pop the max value from the matrix and enter in matches}
matches = \{\}\;
\While{non-zero entry in matrix exists}{
ceiling\textunderscore id, angled\textunderscore id = matrix.argmax()\;
matches[:, angled\textunderscore id] = 0 

\If{ceiling\textunderscore id \textbf{not in} matches}{
matches[ceiling\textunderscore id] = angled\textunderscore id\;
}
}
 \caption{Multi-Camera Tracking algorithm}
\end{algorithm}

We use the homography $H_\text{ceiling$\rightarrow$angled}$ computed in the previous subsection to develop a multi-camera tracking algorithm, by aligning tracks in the \textit{Ceiling} view and the \textit{Angled} view. 

The bounding box from the \textit{Ceiling} view can be treated as a quadrilateral on the pen floor in which the corresponding pig is contained. The bounding box from the \textit{Ceiling} view is then projected under the homography to a quadrilateral in the \textit{Angled} view, as depicted in Figure \ref{fig:homography}. The transformed quadrilateral is then matched with a bounding box in the \textit{Angled} view with which it has significant overlap in terms of the pixel area. As seen in Figure \ref{fig:angled}, ID 56 in the \textit{Ceiling} view will be matched with ID 6 in the \textit{Angled} view.

\section{Evaluation}

\subsection{Camera Deployment}

\begin{figure}[h]
    \centering
    \includegraphics[width=\linewidth]{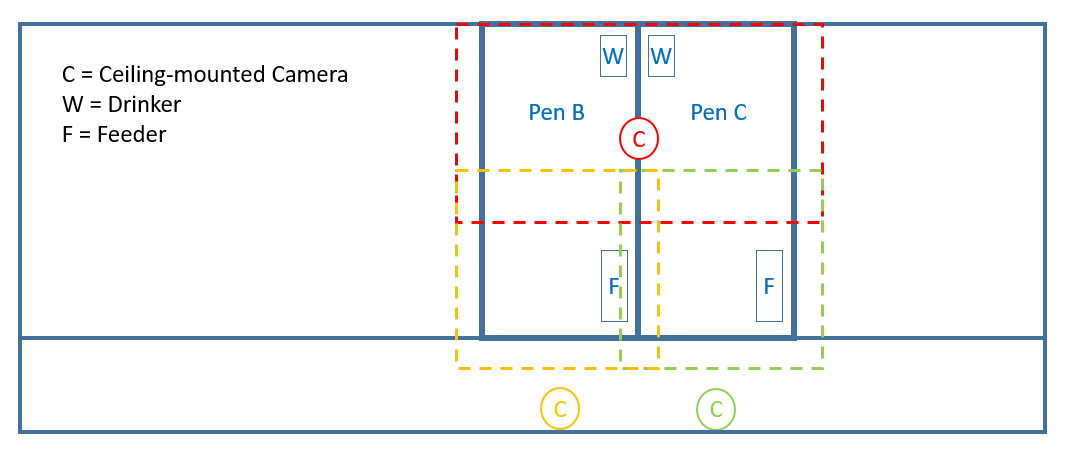}
    \caption{Camera Deployment at the Imported Swine Research Laboratory, UIUC} 
    \label{fig:deployment}
    \vspace{-0.2cm}
\end{figure}

Videos were recorded in two grow-finish pens (B and C) at the University of Illinois, Urbana-Champaign’s Imported Swine Research Laboratory (ISRL), which serves as the research testbed for our experiments. Our surveillance setup monitored 17 and 16 pigs raised in pens B and C respectively. Three wide-angle lens cameras were strategically placed to cover the drinkers and feeders in the visual field, as shown in Figure \ref{fig:deployment}. The \textit{Ceiling} camera covered the drinkers and was placed at a height of 4 metres. The \textit{Angled} cameras covered the feeders and were placed at a height of 2.2 metres. Infrared floodlights were deployed to enhance the existing night vision capabilities of the cameras. The video feed was captured 24x7 at 4k resolution and 15 frames per second. 

\subsection{Dataset}
For training the YOLOv4 model, we sample and annotate 429 images using the VGG Image Annotator \cite{via}. The images were randomly split in an 80:20 between the training set (343 images) and the validation set (86 images). To evaluate the tracking efficacy, five two-minutes long video sequences were annotated with global pig IDs using a custom MATLAB tool. The annotations for two camera views are made every 15\textsuperscript{th} frame, thus (5*2*60*2) 1200 annotated frames are available, in addition to the 429 images for object detection.

\subsection{YOLOv4 detection}
We train the YOLOv4 model on an Nvidia V100 GPU provided by the HAL computing cluster \cite{hal}. 
The model was configured for an input resolution of 608x608 and optimized for 2000 iterations with a batch size of 64, a learning rate of 0.001, and momentum of 0.95. We evaluate the model using standard object detection metrics. A mean Average Precision of 99.5\% and an average Intersection over Union of 80.52\% is achieved on the validation set.

\subsection{Tracking evaluation}
We evaluate the efficacy of our tracking algorithm using standard Multi-Object Tracking Metrics \cite{ristani2016performance}.  The Multi-Object Tracking Accuracy (MOTA) takes into account three sources of errors and can be defined by the following equation: MOTA = 1 - $ \frac{\sum_{t}(FN_{t} + FP_{t} + IDSW_{t})} {\sum\limits_{t} GT_{t}}$, where $FN_t$, $FP_t$ and $IDSW_t$ are False Negatives, False Positives and Identity Switches, and $GT_t$ is the ground truth number of bounding boxes at time at time $t$. Multi-Object Tracking Precision (MOTP) captures the localization precision of the detector. Standard metrics such as Precision and Recall measure the number of mismatched or unmatched detection-frames, regardless of where the discrepancies start or end or which cameras are involved. We also report identity related metrics computed after global min-cost matching: IDF1, IDP, and IDR. IDF1 is the ratio of correctly identified detections over the average number of ground-truth and computed detections, IDP (precision) is the fraction of computed detections that are correct whereas IDR (recall) accounts for correctly identified ground truth detections. 

We evaluate our multi-camera tracking approach by reporting the Camera Handover Accuracy (CHA). CHA is defined as the fraction of predicted identity matches for two camera views out of the total ground truth matches. Intuitively, MOTA and MOTP quantifies how accurately the algorithm is tracking pigs with minimal false positives, and CHA quantifies how accurately identities are exchanged between cameras using homography. As seen in Table \ref{tab:metrics}, we observe that tracking is relatively easier during day as the quality of video feed degrades at night. We achieve an MOTA, MOTP and CHA of 65\%, 54.3\% and 74\% respectively. 

\section{Discussion}

\subsection{Issues in Detecting Pigs using YOLOv4}
\begin{figure}[!ht]
\centering

\includegraphics[width=\linewidth]{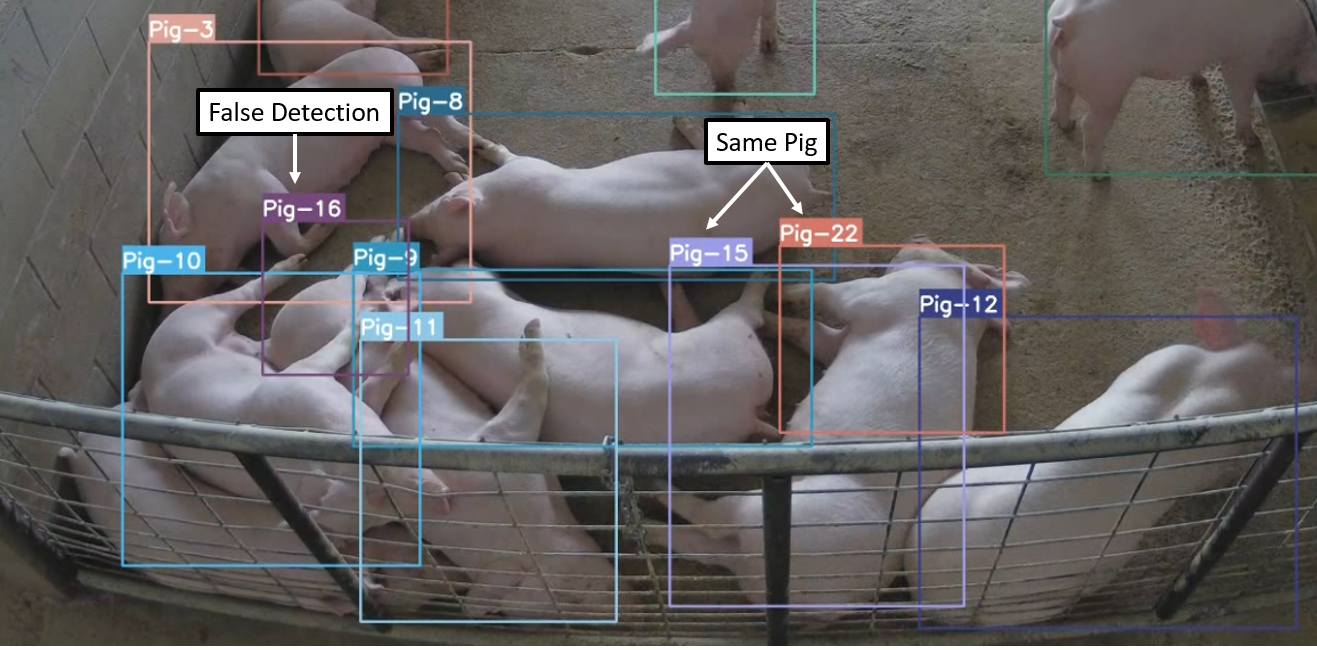}
\caption{False Positives and False Negatives in Detection} 
\label{fig:huddling}
\vspace{-0.2cm}
\end{figure}

Occlusions are a primary source of false negatives. Even though detecting pigs by a direct application of state-of-the-art deep learning models, such as YOLOv4, yield great results, there are several scenarios where detection of pigs becomes very challenging. One of the most common impediments to detection is occlusion. If a pig is partially or completely occluded in any of the views, then the model tends to either miss the pig completely or predicts a bigger bounding box, encompassing more than one pig, which leads to missed detections. Having two camera views helps in partially alleviating this problem. 

False positives result from part of pig being recognized as a whole pig. Pigs tend to huddle in pens to keep themselves warm as a group, which creates a challenge for recognition. Usually, a post-processing non-maximal suppression step in the object detection pipeline helps in selecting the best bounding box with the maximum confidence score. But often, the model considers a part of the pig to be the whole pig and predicts a box that is not suppressed during non-maximal suppression. 

One can see both kinds of errors, as described above, in Figure \ref{fig:huddling}. The pig in the bottom left of the frame is not detected by the model due to heavy occlusion from other pigs and the pen door. 
Additionally, detections `Pig-22' and `Pig-15' in Figure \ref{fig:huddling} belong to the same pig. Here, `Pig-22' predicts a bounding box for a part of a pig and should be ideally suppressed. It is also worth noting that `Pig-16' is a false detection as the two hind legs of two different pigs confuse the model into believing that it is a single pig.      

\subsection{Issues in Tracking Pigs using DeepSORT}

\begin{figure}[h]
    \centering
    \includegraphics[width=\linewidth]{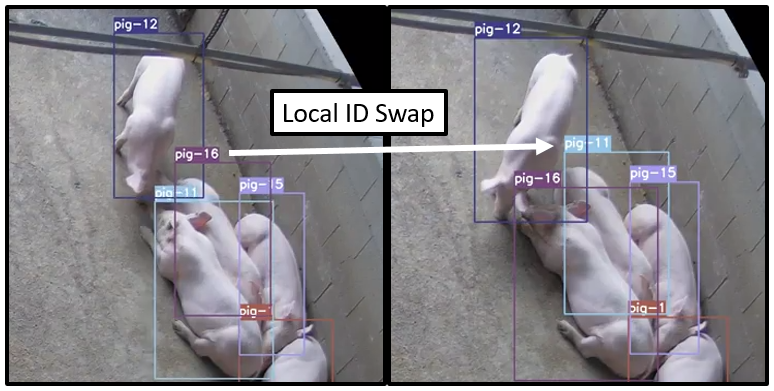}
    \caption{Identity Swap in DeepSORT} 
    \label{fig:id-swap}
    \vspace{-0.2cm}
\end{figure}

The detection issues described in the previous subsection affect the tracking performance of DeepSORT, as false positives lead to an extra number of tracks and false negatives lead to premature termination of tracks. Another primary source of tracking errors is induced by rapid movements in a pig huddle. As seen in Figure \ref{fig:id-swap}, the identity of Pigs 11 and 16 get swapped due to a rapid social interaction with Pig 12. Theoretically, DeepSORT should tackle and minimize such identity swaps as it uses an appearance model for re-identification. But the lack of visually distinguishing features between pigs calls for advanced re-identification algorithms while tracking. One potential solution might be to apply another recognition approach, such as a gait analysis, to reidentify them upon occlusion.

\subsection{Issues in Global Tracking}
The inherent problems in detection and tracking described in the previous subsections affect global tracking, as false detections can lead to false correspondences between different views. The algorithm is also sensitive to variations in detections from a pig huddle, as any error in homography estimation can magnify and lead to an identity swap between multiple pigs. This issue can be mitigated by precise homography estimation and by only assigning a global ID for pigs that have higher overlapping confidence. But that can lead to missed opportunities of matching, leading to a low recall value. Additionally, the multi-camera algorithm described in section \ref{sec:global-tracking} works under the assumption that all the cameras are synchronized with each other. But in actual deployments, there is a variable delay of several minutes between the capture times of the \textit{Ceiling} and the \textit{Angled} views. The delay needs to be fixed manually as such homography-based multi-camera trackers need tight synchronization between the two views. 

\section{Conclusion}

Livestock monitoring is becoming increasingly important in precision livestock management. In order to tackle the problem of monitoring livestock residing in large pens, we propose a scheme for homography-based multi-camera tracking. We achieve a tracking accuracy and precision of 66.7\% and 76.2\%, respectively. We plan to further enrich our multi-camera tracking dataset with key focal points and a validated ethogram for pig behavior. We open-source this dataset so that it can be leveraged to build educational applications for pig tracking and behavior monitoring.

{\small
\bibliographystyle{ieee_fullname}
\bibliography{egbib}
}

\end{document}